# Variational Gaussian Dropout is not Bayesian

**Jiri Hron**
University of Cambridge
jh2084@cam.ac.uk

**Alexander G. de G. Matthews**
University of Cambridge
am554@cam.ac.uk

**Zoubin Ghahramani**
University of Cambridge, UBER AI Labs
zoubin@eng.cam.ac.uk

## Abstract

Gaussian multiplicative noise is commonly used as a stochastic regularisation technique in training of deterministic neural networks [12]. A recent paper [6] reinterpreted the technique as a specific algorithm for approximate inference in Bayesian neural networks; several extensions ensued [8, 10, 11]. We show that the log-uniform prior used in all the above publications does not generally induce a proper posterior, and thus Bayesian inference in such models is ill-posed. Independent of the log-uniform prior, the correlated weight noise approximation proffered in [6] has further issues leading to either infinite objective or high risk of overfitting. The above implies that the reported sparsity of obtained solutions cannot be explained by Bayesian or the related minimum description length arguments. We thus study the objective from a non-Bayesian perspective, provide its previously unknown analytical form which allows exact gradient evaluation, and show that the reparametrisation proposed in [10] introduces minima not present in the original [6]. Implications and future research directions are discussed.

## 1 Introduction

Variational inference (VI) approximates Bayesian posterior distribution over a set of latent variables $W$ by optimising the evidence lower bound (ELBO) $\mathcal{L}(q) = \mathbb{E}_{q(W)}[\log p(Y \mid X, W)] - \mathrm{KL}(q(W) \| p(W))$ with respect to an approximate posterior $q(W)$. Variational dropout [6] attempts to approximate a Bayesian neural network (BNN) by a posterior factorised over individual weights $w \in W$, $q(w) = \mathcal{N}(\theta, \alpha\theta^2)$. This corresponds to multiplying the individual mean parameters $\theta \in \boldsymbol{\theta}$ by $\varepsilon \sim \mathcal{N}(1, \alpha)$ as in the Gaussian multiplicative dropout [12]. The prior factorises in the same way and is chosen so that $\mathrm{KL}(q(W) \| p(W))$ is independent of the mean parameters $\boldsymbol{\theta}$.

In Section 2, we show that the chosen improper log-uniform prior [6] generally does not induce a proper posterior, and thus the reported sparsification [10] cannot be explained by the standard Bayesian and the related minimum description length (MDL) arguments. In this sense, Variational Dropout falls into the same category of non-Bayesian sparsification approaches as, for example, Lasso [13], because the uncertainty estimates based on $q(W)$ do not have the usual interpretation, and the model may exhibit overfitting. The danger of overfitting might also be high when using the correlated weight noise parametrisation [6] as discussed in Section 3. Consequently, we study the objective in Section 4, derive its previously unknown analytical form which allows us to obtain exact gradients in a computationally efficient way, and prove that it favours posterior approximations $\mathcal{N}(\theta, \alpha\theta^2)$ with high variance relative to the value of mean $\theta$ (i.e. with $\alpha \gg 0$). We further show that the parametrisation of the approximate posterior proposed in [10] is not equivalent to the original one [6], and discuss the implications which can partially explain the reported sparsification.

## 2  Improper prior and posterior

The log-uniform prior proposed in [6] factorises over $w \in \boldsymbol{W}$ with $\mathrm{p}(w) := \mathrm{C}/|w|$, which is equivalent to putting a uniform prior on $\log|w|$. This is an improper prior which means that there is no $\mathrm{C} \in \mathbb{R}$ for which $\mathrm{p}(w)$ is a valid probability density. Nevertheless, improper priors can sometime lead to proper posteriors (e.g. normal Jeffreys prior for Gaussian likelihood with unknown mean and variance parameters) if C is treated as a positive finite constant.

For any proper posterior distribution, the normaliser $\mathrm{Z} = \int_{\mathbb{R}^\mathrm{D}} \mathrm{d}(\boldsymbol{W}) \mathrm{p}(\boldsymbol{Y} \,|\, \boldsymbol{X}, \boldsymbol{W}) \mathrm{p}(\boldsymbol{W})$ has to be finite (D denotes the total number of weights). We will now show that this requirement is generally not satisfied for the log-uniform prior combined with commonly used likelihood functions.

**Proposition 1.** *Assume the log-uniform prior is used and that there exists some $w \in \boldsymbol{W}$ such that the likelihood function at $w = 0$ is continuous in $w$ and non-zero. Then the posterior is improper.*

All proofs can be found in the appendix. Notice that standard architectures with activations like rectified linear or sigmoid, and Gaussian or Categorical likelihood satisfy the above assumptions, and thus the posterior distribution for non-degenerate datasets will generally be improper.

Furthermore, the pathologies are not limited to the region near $w = 0$, but can also arise in the tails. As an example, we will consider a single variable Bayesian logistic regression problem $\mathrm{p}(y \,|\, x, w) = 1/(1 + \exp(-xw))$, and again use the log-uniform prior for $w$. For simplicity, assume that we have seen a single observation $(x = 1, y = 1)$ and wish to infer the posterior distribution. To show that the right tail has infinite mass, we integrate over intervals of the form $[k, \infty), k > 0$,

$$\int_{[k,\infty)} \mathrm{d}(w) \mathrm{p}(w) \mathrm{p}(y \,|\, x, w) = \int_{[k,\infty)} \mathrm{d}(w) \frac{\mathrm{C}}{|w|} \frac{1}{1 + \exp(-w)} > \int_{[k,\infty)} \mathrm{d}(w) \frac{\mathrm{C}}{|w|} \frac{1}{1 + \exp(-k)}$$

$$= \frac{\mathrm{C}}{1 + \exp(-k)} \log|w| \Big|_k^\infty = \frac{\mathrm{C}}{1 + \exp(-k)} \cdot (\infty - \log k) = \infty\,.$$

Equivalently, we could have obtained infinite mass in the left tail – for example by taking the observation to be $(x = -1, y = 1)$. Because the sigmoid function is continuous and equal to $1/2$ at $w = 0$, the posterior also has infinite mass around the origin, exemplifying both of the discussed degeneracies. The normalising constant is obviously infinite and thus the posterior is again improper.

In general, improper posteriors lead to undefined or incoherent inferences. The above shows that this is the case for the log-uniform prior combined with BNN and related models, making Bayesian inference, exact or approximate, in such models ill-posed.

## 3  Support mismatch in correlated Variational dropout

Standard Bernoulli dropout multiplies each input to a layer by a binary random variable which is equivalent to multiplying the corresponding row of the subsequent weight matrix by the same variable. To emulate this behaviour, Section 3.2 in [6] proposes to use a shared Gaussian random variable for whole rows of the posterior weight matrices, $\boldsymbol{w}_i = s_i \boldsymbol{\theta}_i, s_i \sim \mathcal{N}(1, \alpha)$. This leads to a degenerate posterior which only assigns mass along the directions defined by $\boldsymbol{\theta}$.

There are two possible consequences. If the log-uniform prior is used, then the directions defined by $\boldsymbol{\theta}$ span a measure zero subspace of $\mathbb{R}^\mathrm{D}$ and thus the $\mathrm{KL}\,(\mathrm{q}(\boldsymbol{W}) \| \mathrm{p}(\boldsymbol{W}))$ and consequently $\mathrm{KL}\,(\mathrm{q}(\boldsymbol{W}) \| \mathrm{p}(\boldsymbol{W} \,|\, \boldsymbol{X}, \boldsymbol{Y}))$ are equal to infinity for any configuration of $\boldsymbol{\theta}$ (see, for example, [9, Section 2.1]). This means that there is no Bayesian interpretation for the obtained $\mathrm{q}(\boldsymbol{W})$. If the prior is instead only defined over the $s_i$ scalars, $\boldsymbol{\theta}$ become model parameters and thus not regularised. Optimising the ELBO will lead to a valid Bayesian posterior approximation for the $s_i$ variables; however, the large number of unregularised parameters $\boldsymbol{\theta}$ can potentially lead to significant overfitting. Both interpretations of the correlated weight noise approximation are thus problematic.

## 4  Variational dropout as pseudo KL divergence minimisation

We have established that optimisation of the ELBO implied by a BNN with log-uniform prior over its weights cannot generally be interpreted as a form of approximate Bayesian inference. Nevertheless, the reported empirical results suggest that the objective might possess reasonable properties resulting in sparse solutions. We present some preliminary observations and related literature below, leaving more thorough exploration of the topic to future research. In the following paragraphs, we prepend all terms related to VI terminology by *pseudo* to emphasize that the standard properties need not hold.

Firstly, despite the contradictory claims in [6, 8, 10], there exists an analytical expression for the pseudo KL $(q(w)\| p(w))$, and for its derivative with respect to the pseudo variational parameters.

**Remark 1.** *Consider* $q(w) = \mathcal{N}(\mu, \sigma^2)$, *and* $p(w) = C/|w|$, *and denote* $u := \mu^2/(2\sigma^2) \geq 0$. *Then,*

$$\mathrm{KL}\left(q(w)\| p(w)\right) = -\frac{1}{2}\log(2\pi e) - \log C + \frac{1}{2}\left(\log 2 + e^{-u}\sum_{k=0}^{\infty}\frac{u^k}{k!}\psi(1/2 + k)\right) \quad (1)$$

$$= -\frac{1}{2}\log(\pi e) - \log C - \frac{1}{2}\left(\left.\frac{\partial}{\partial a}M(a; 1/2; -u)\right|_{a=0} + \gamma + 2\log 2\right), \quad (2)$$

*where* $\psi(x)$ *denotes the digamma function, and* $M(a; b; z)$ *the Kummer's function of the first kind. We can obtain gradients with respect to* $\mu$ *and* $\sigma^2$ *using,*

$$\nabla_u \mathrm{KL}\left(q(w)\| p(w)\right) = \begin{cases} 1 & u = 0 \\ \frac{D_+(\sqrt{u})}{\sqrt{u}} & u > 0 \end{cases}, \quad (3)$$

*and the chain rule;* $D_+(x)$ *is the Dawson integral. The derivative is continuous in* $u$ *on* $[0, \infty)$.

Equation (3) is sufficient for first order gradient-based pseudo ELBO optimisation, and thus can be used to replace the approximations used in [6, 10]. We note that numerically accurate implementations of the Dawson integral exist in many popular programming languages (see, for example, [5]).

In variational inference literature, the term $\mathrm{KL}(q(w)\| p(w))$ is sometimes interpreted as a regulariser, constraining $q(w)$ from concentrating at the maximum likelihood estimate which would be optimal with respect to the other term $\mathbb{E}_{q(W)}[\log p(Y \mid X, W)]$ in the ELBO. It is thus natural to ask what effect this term has on the pseudo variational parameters. Noticing that only the infinite sum in Equation (1) depends on these parameters, and that the first summand is always equal to $\psi(1/2)$, we can focus on summands corresponding to $k \geq 1$. Because $\psi(1/2 + k) > 0, \forall k \geq 1$, all summands are non-negative. Hence the penalty will be minimised if $\mu^2/(2\sigma^2) = 0$, i.e. when $\mu = 0$ and/or $\sigma^2 \to \infty$; Remark 2 is sufficient to establish that this minimum is unique.

**Remark 2.** *Under Remark 1 assumptions,* $\mathrm{KL}(q(w)\| p(w))$ *is strictly increasing for* $u \in [0, \infty)$.

In the multiplicative parametrisation [6], $u = \mu^2/(2\sigma^2) = \theta^2/(\alpha\theta^2) = 1/\alpha$ and thus the optimum can only be approached as $\alpha \to \infty$. Hence the additive parametrisation proposed in [10] is not equivalent to the multiplicative, which combined with the improved approximation of the $\mathrm{KL}(q(w)\| p(w))$ term might explain why the original paper [6] did not report the same sparsification as [10].

Section 2 suggests the pathological behaviour is non-trivial to remove when log-uniform prior is used. Some proper prior distributions might, however, lead to comparable sparsification [8]. Alternatively, Remarks 1 and 2 imply that pseudo ELBO optimisation leads to penalised maximum likelihood estimation of the pseudo variational parameters. Noting that standard differential entropy can be interpreted as quantifying pseudo KL from the D-dimensional Lebesgue measure, it might be interesting to ask what can be gained by replacing it by the log-uniform or other non-finite measure. A starting point might be some of the previous applications of entropy for regularisation [7, 16, 17].

## 5 Conclusion

Bayesian and the related MDL interpretations of the Variational Gaussian Dropout are technically flawed, and thus cannot be used to explain the reported results. We studied the pseudo VI objective, derived its analytical form, and provided a simple way to optimise it without resorting to any further approximations. We remarked that replacing the multiplicative by additive posterior parametrisation does affect characteristics of the objective, showing that the change may partially explain why [6] did not report the same sparsification as [10], and outlined some interesting future research directions.

## References

[1] G. Dattoli and H. Srivastava. A Note on Harmonic Numbers, Umbral Calculus and Generating Functions. *Applied Mathematics Letters*, 21(7):686–693, 2008.

[2] R. W. Gosper. Harmonic Summation and Exponential GFS, 1996.

[3] T. Gowers. Differentiating Power Series. https://gowers.wordpress.com/2014/02/22/differentiating-power-series/, February 2014.

[4] F. E. Harris. Tables of the Exponential Integral Ei(x). *Mathematical Tables and Other Aids to Computation*, 11(57):9–16, 1957.

[5] S. G. Johnson. Faddeeva Package. http://ab-initio.mit.edu/wiki/index.php/Faddeeva_Package.

[6] D. P. Kingma, T. Salimans, and M. Welling. Variational Dropout and the Local Reparameterization Trick. In C. Cortes, N. D. Lawrence, D. D. Lee, M. Sugiyama, and R. Garnett, editors, *Advances in Neural Information Processing Systems 28*, pages 2575–2583. Curran Associates, Inc., 2015.

[7] V. Koltchinskii. Sparse Recovery in Convex Hulls via Entropy Penalization. *The Annals of Statistics*, pages 1332–1359, 2009.

[8] C. Louizos, K. Ullrich, and M. Welling. Bayesian Compression for Deep Learning. *arXiv preprint arXiv:1705.08665*, 2017. To be published in the proceedings of NIPS 2017.

[9] A. G. de G. Matthews. *Scalable Gaussian process inference using variational methods*. PhD thesis.

[10] D. Molchanov, A. Ashukha, and D. Vetrov. Variational Dropout Sparsifies Deep Neural Networks. In *Proceedings of the 34th International Conference on Machine Learning*, volume 70 of *Proceedings of Machine Learning Research*, pages 2498–2507. PMLR, 2017.

[11] K. Neklyudov, D. Molchanov, A. Ashukha, and D. Vetrov. Structured Bayesian Pruning via Log-Normal Multiplicative Noise. *arXiv preprint arXiv:1705.07283*, 2017. To be published in the proceedings of NIPS 2017.

[12] N. Srivastava, G. E. Hinton, A. Krizhevsky, I. Sutskever, and R. Salakhutdinov. Dropout: A Simple Way to Prevent Neural Networks from Overfitting. *Journal of Machine Learning Research*, 15(1):1929–1958, 2014.

[13] R. Tibshirani. Regression Shrinkage and Selection via the Lasso. *Journal of the Royal Statistical Society. Series B (Methodological)*, pages 267–288, 1996.

[14] Wolfram|Alpha. https://goo.gl/A5bxLh, November 2017.

[15] Wolfram|Alpha. https://goo.gl/sZoiuC, November 2017.

[16] T. Zhang. Regularized Winnow Methods. In T. K. Leen, T. G. Dietterich, and V. Tresp, editors, *Advances in Neural Information Processing Systems 13*, pages 703–709. MIT Press, 2001.

[17] T. Zhang et al. From epsilon-Entropy to KL-Entropy: Analysis of Minimum Information Complexity Density Estimation. *The Annals of Statistics*, 34(5):2180–2210, 2006.

## A  Proofs

*The following notation and identities are used throughout this section:* $\psi(x)$ is the digamma function, $\psi(x+1) = \psi(x) + 1/x$, $\psi(k+1) = H_k - \gamma$ where $H_k$ is the $k^{th}$ harmonic number and $\gamma$ is the Euler–Mascheroni's constant, $\text{Ei}(x) = -\int_{-x}^{\infty} d(t) e^{-t}/t$ is the exponential integral function, $\sum_{k=1}^{\infty} u^k H_k / k! = e^u (\gamma + \log|u| - \text{Ei}(-u))$ [1, 2], and $\sum_{k=1}^{\infty} u^k/(k! k) = \text{Ei}(u) - \gamma - \log|u|$ [4]; the last two identities hold for $u > 0$. Importantly, we define $0^0 := 1$ unless stated otherwise.

*Proof of Proposition 1.* Denote the likelihood value by $\epsilon > 0$. Take an arbitrary number $r$ such that $\epsilon > r > 0$. By continuity, we can find $\delta > 0$ such that $|w - 0| < \delta$ implies that the likelihood value is greater than $r$; let $A \ni 0$ denote the open ball of radius $\delta$ centred at $0$. Because both the prior and the likelihood only take non-negative values, we have,

$$Z = \int_{\mathbb{R}^{D-1}} \mathrm{d}(\boldsymbol{W} \setminus w)\mathrm{p}(\boldsymbol{W} \setminus w) \left[ \int_{\mathbb{R}} \mathrm{d}(w)\mathrm{p}(w)\mathrm{p}(\boldsymbol{Y} \mid \boldsymbol{X}, \boldsymbol{W}) \right]$$
$$> \int_{\mathbb{R}^{D-1}} \mathrm{d}(\boldsymbol{W} \setminus w)\mathrm{p}(\boldsymbol{W} \setminus w) \left[ \int_A \mathrm{d}(w) \frac{\mathrm{C}}{|w|} r = \mathrm{C} r \cdot \infty \right] = \infty \,.$$

$Z = \infty$ means that the measure of $\mathbb{R}^D$ under $\mathrm{p}(\boldsymbol{W} \mid \boldsymbol{X}, \boldsymbol{Y})$ is equal to infinity, and thus it cannot be renormalised into a proper probability density. $\square$

*Proof of Remark 1.* Using standard identities about Gaussian random variables, and the fact that $v := \varepsilon^2$, $\varepsilon \sim \mathcal{N}(\mu/\sigma, 1)$, follows the non-central chi-squared distribution $\chi^2(\lambda, \nu)$ with $\nu = 1$ degrees of freedom and non-centrality parameter $\lambda = (\mu/\sigma)^2$, we have,

$$\mathbb{E}_{\mathrm{q}(w)}[\log \mathrm{q}(w)] - \mathbb{E}_{\mathrm{q}(w)}[\log \mathrm{p}(w)] = \mathbb{E}_{\mathrm{q}(w)}[\log \mathrm{q}(w)] - \log \mathrm{C} + \frac{1}{2} \mathbb{E}_{\mathrm{q}(w)}[\log|w|^2]$$
$$= -\frac{1}{2}\log(2\pi\mathrm{e}\sigma^2) - \log \mathrm{C} + \frac{1}{2} \mathbb{E}_{\varepsilon \sim \mathcal{N}(\mu/\sigma,1)}[\log \sigma^2 \varepsilon^2]$$
$$= -\frac{1}{2}\log(2\pi\mathrm{e}\sigma^2) - \log \mathrm{C} + \frac{1}{2}\left(\log \sigma^2 + \mathbb{E}_{v \sim \chi^2(\mu^2/\sigma^2,1)}[\log v]\right)$$
$$= -\frac{1}{2}\log(2\pi\mathrm{e}\sigma^2) - \log \mathrm{C} + \frac{1}{2}\left(\log \sigma^2 + \sum_{k=0}^{\infty} \mathrm{e}^{-\frac{\mu^2}{2\sigma^2}} \frac{(\frac{\mu^2}{2\sigma^2})^k}{k!} \mathbb{E}_{v \sim \chi^2(0,1+2k)}[\log v]\right)$$
$$= -\frac{1}{2}\log(2\pi\mathrm{e}) - \log \mathrm{C} + \frac{1}{2}\left(\log 2 + \mathrm{e}^{-\frac{\mu^2}{2\sigma^2}} \sum_{k=0}^{\infty} \frac{(\frac{\mu^2}{2\sigma^2})^k}{k!} \psi(1/2 + k)\right)$$
$$= -\frac{1}{2}\log(\pi\mathrm{e}) - \log \mathrm{C} - \frac{1}{2}\left( \left.\frac{\partial}{\partial a}\mathrm{M}(a; 1/2; -\mu^2/(2\sigma^2))\right|_{a=0} + \gamma + 2\log 2\right),$$

where $\mathrm{M}(a; b; z)$ denotes the Kummer's function of the first kind. We used the fact that the density function of $\chi^2(\lambda, \nu)$ can be written in terms of a Poisson mixture of centralised chi-squared distributions, and the identity $\mathbb{E}_{v \sim \chi^2(0,\nu)}[\log v] = \psi(\nu/2) - \log(1/2)$. It is easy to check that Equation (1) holds for all $u \geq 0$ as long as we define $0^0 = 1$, and keep $0^k = 0, \forall k > 0$. The last equality was obtained using Wolfram Alpha [14]; to validate this result, we performed an extensive numerical test, and will now show that the series indeed converges for $u = \mu^2/(2\sigma^2) \in [0, \infty)$, i.e. for all plausible values of $u$. The comparison test gives us convergence for $u \in (0, \infty)$:

$$\sum_{k=0}^{\infty} \frac{u^k}{k!} \psi(1/2 + k) < \psi(1/2) + \sum_{k=1}^{\infty} \frac{u^k}{k!} \psi(1 + k) = \psi(1/2) + \sum_{k=1}^{\infty} \frac{u^k}{k!} (\mathrm{H}_k - \gamma)$$
$$= \psi(1/2) + \mathrm{e}^u(\gamma + \log|u| - \mathrm{Ei}(-u)) - \gamma(\mathrm{e}^u - 1) = \psi(1/2) - \gamma + \mathrm{e}^u(\log|u| - \mathrm{Ei}(-u)),$$

where we use the fact that the individual summands are non-negative for $k \geq 1$ (which is also means we need not take the absolute value explicitly). It is trivial to check that the series converges at $u = 0$, and thus we have convergence for all $u \in [0, \infty)$.

To obtain the derivative with respect to $u$, we use the infinite series formulation from Equation (1), and the fact that the derivative of a power series within its radius of convergence is equal to the sum of its term-by-term derivatives (see [3] for a nice proof which avoids complex analysis). Using the fact that only the infinite series in Equation (1) depends on $u$, we obtain,

$$\nabla_u e^{-u} \sum_{k=0}^{\infty} \frac{u^k}{k!} \psi(1/2+k) = \nabla_u \left( e^{-u} \psi(1/2) + e^{-u} \sum_{k=1}^{\infty} \frac{u^k}{k!} \psi(1/2+k) \right)$$

$$= -e^{-u} \psi(1/2) + e^{-u} \sum_{k=1}^{\infty} \left( \frac{u^{k-1}}{(k-1)!} \psi(1/2+k) \right) - e^{-u} \sum_{k=1}^{\infty} \left( \frac{u^k}{k!} \psi(1/2+k) \right)$$

$$= e^{-u} (\psi(3/2) - \psi(1/2)) + e^{-u} \sum_{k=1}^{\infty} \left( \frac{u^k}{k!} \psi(3/2+k) \right) - e^{-u} \sum_{k=1}^{\infty} \left( \frac{u^k}{k!} \psi(1/2+k) \right)$$

$$= 2e^{-u} + e^{-u} \sum_{k=1}^{\infty} \frac{u^k}{k!} \frac{1}{1/2+k} = e^{-u} \sum_{k=0}^{\infty} \frac{u^k}{k!} \frac{1}{1/2+k} = \frac{2D_+(\sqrt{u})}{\sqrt{u}},$$

for $u > 0$ and is equal to 2 if $u = 0$; in our case, the condition $u \geq 0$ is satisfied by definition; to obtain the expression in Equation (3), notice that the above series is multiplied by $1/2$ in Equation (1). Equality of the last infinite series to $2D_+(\sqrt{u})/\sqrt{u}$, was again obtained using Wolfram Alpha [15]; the result was numerically validated, and convergence on $u \in (0, \infty)$ can again be established using the comparison test:

$$\sum_{k=0}^{\infty} \left| \frac{u^k}{k!} \frac{1}{1/2+k} \right| = \sum_{k=0}^{\infty} \frac{u^k}{k!} \frac{1}{1/2+k} < 2 + \sum_{k=1}^{\infty} \frac{u^k}{k!} \frac{1}{k} = 2 + \text{Ei}(u) - \gamma - \log|u| \ .$$

The convergence at $u = 0$ can be checked trivially, yielding convergence for all $u \in [0, \infty)$.

$D_+(u)$ and $\sqrt{u}$ are continuous on $(0, \infty)$, and $\sqrt{u} > 0$; hence $D_+(u)/\sqrt{u}$ is continuous on $(0, \infty)$, and we observe $\lim_{u \to 0_+} D_+(\sqrt{u})/\sqrt{u} = 1$, i.e. the gradient is continuous in $u$ on $[0, \infty)$. □

*Proof of Remark 2.* We use the conclusion of Remark 1 which established that the pseudo KL is differentiable for $u \in [0, \infty)$ (and thus continuous on the same interval). To show that $\text{KL}(q(w) \| p(w))$ is strictly increasing for $u \in [0, \infty)$, it is sufficient to observe,

$$\nabla_u \text{KL}(q(w) \| p(w)) = \frac{1}{2} e^{-u} \sum_{k=0}^{\infty} \frac{u^k}{k!} \frac{1}{1/2+k} > 0 \,,$$

because each summand is strictly positive for $u \in [0, \infty)$ (given $0^0 = 1$). By a simple application of the mean value theorem, we conclude $\text{KL}(q(w) \| p(w))$ is strictly increasing in $u$ on $[0, \infty)$. □